\documentclass[conference]{IEEEtran}
\IEEEoverridecommandlockouts
\usepackage{cite}
\usepackage{amsmath,amssymb,amsfonts}
\usepackage{algorithmic}
\usepackage{graphicx}
\usepackage{textcomp}
\usepackage{xcolor}
\usepackage{times}
\usepackage{soul}
\usepackage{url}
\usepackage[utf8]{inputenc}
\usepackage{graphicx}
\usepackage{amsmath}
\usepackage{amsthm}
\usepackage{booktabs}
\usepackage{algorithm}
\usepackage[switch]{lineno}

\usepackage{subcaption}
\usepackage{multirow}
\usepackage[misc]{ifsym} 

\def\BibTeX{{\rm B\kern-.05em{\sc i\kern-.025em b}\kern-.08em
    T\kern-.1667em\lower.7ex\hbox{E}\kern-.125emX}}
\begin{document}

\title{Elastic Architecture Search 
for Efficient Language Models}


\author{\IEEEauthorblockN{Shang Wang}
\IEEEauthorblockA{
Indepentdent Researcher\\
email.shangwang@gmail.com
}
}

\maketitle

\begin{abstract}
As large pre-trained language models become increasingly critical to natural language understanding (NLU) tasks, their substantial computational and memory requirements have raised significant economic and environmental concerns. Addressing these challenges, this paper introduces the Elastic Language Model (ELM), a novel neural architecture search (NAS) method optimized for compact language models. ELM extends existing NAS approaches by introducing a flexible search space with efficient transformer blocks and dynamic modules for dimension and head number adjustment. These innovations enhance the efficiency and flexibility of the search process, which facilitates more thorough and effective exploration of model architectures. We also introduce novel knowledge distillation losses that preserve the unique characteristics of each block, in order to improve the discrimination between architectural choices during the search process. Experiments on masked language modeling and causal language modeling tasks demonstrate that models discovered by ELM significantly outperform existing methods.
\end{abstract}

\begin{IEEEkeywords}
Language Models, Natural Language Processing, Neural Architecture Search
\end{IEEEkeywords}

\section{Introduction}
Natural language understanding (NLU) is a critical component of natural language processing (NLP). Recent advancements in NLU have been driven by large pretrained language models \cite{devlin2019bert, liu2019roberta, radford2019language, touvron2023llama}, which have significantly improved performance across various tasks by expanding model size and data scale. However, these large models come with challenges, including high computational and memory requirements that make them costly to train and deploy. Such demands not only restrict their accessibility but also heighten environmental concerns due to considerable energy consumption.

To mitigate these challenges, researchers have turned to neural architecture search (NAS) to develop more efficient BERT models. However, most efforts have focused on optimizing moderate-sized models like BERT-tiny (with approximately 15M parameters). When attempting to further reduce model size, a notable performance drop occurs. This decline is primarily attributed to the limited flexibility of existing NAS methods, which operate within a restricted search space with fixed block types, dimensions, and head numbers. This rigidity restricts the ability to finely tune architectures, which is essential for achieving optimal performance in smaller models.

To address this issue, in this paper, we introduce ELM, a novel NAS method designed to search for flexible and efficient language models. ELM expands the search space to include both BERT ~\cite{devlin2019bert} and MobileBERT ~\cite{sun2020mobilebert} blocks, enhanced with additional weight-sharing options in attention mechanisms. To further enhance the flexibility of our approach, we propose a dynamic dimension search strategy, which is guided by the scores from principal component analysis (PCA) observed during training. This approach allows for the adaptive adjustment of block dimensions based on the evolving importance of features throughout training. Additionally, we employ centered kernel alignment (CKA) to evaluate similarities between attention heads, which facilitates the fusion of redundant heads to improve model efficiency. With these techniques, ELM enables more effective training and better utilization of critical features.

Furthermore, we introduce a novel approach to knowledge distillation (KD), inspired by DIST \cite{huang2022knowledge}, to overcome the limitations of traditional KD methods. The proposed KD approach uses relational losses that preserve the diversity and discriminability of different blocks during the search process, leading to a more accurate evaluation of architecture performance.

Our contributions are threefold:
\begin{enumerate}
    \item We introduce a flexible search space for language models that incorporates efficient candidate blocks. We further propose dynamic strategies for dimension and head number adjustments, which is informed by PCA and CKA.
    \item We address the shortcomings of traditional KD methods by introducing relational losses in our knowledge distillation approach. This helps maintain the unique characteristics and diversity of architectural blocks, and thereby enhancing the effectiveness of the NAS process.
    \item Through comprehensive experiments on masked language modeling and causal language modeling tasks, we demonstrate the superiority of our approach, which achieves significant advancements over existing methods.
\end{enumerate}

\begin{figure*}[t]
\centering
\includegraphics[scale=0.48]{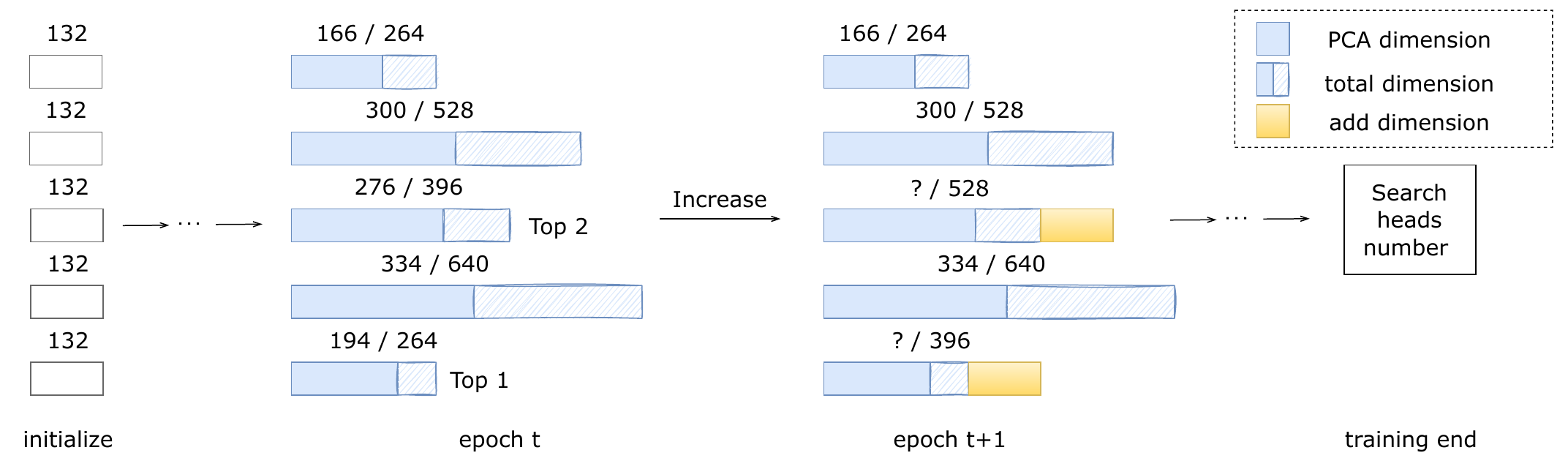}
\begin{center}
\vspace{-3mm}
\captionsetup{singlelinecheck=off}
\caption{The diagram depicts the expansion of the hidden dimension wihtin the feed-forward network (FFN), using five blocks as an example. Initially, the dimension of each block is set to 132. After each epoch, the dimensions of the top two blocks with the highest PCA values are increased by 132. It is crucial to note that the dimensions corresponding to fixed PCA values also change after their increase, which is indicated by ``?'' in the diagram.}
\label{fig:search_space}
\end{center}
\vspace{-3mm}
\end{figure*}

\section{Related Work}

\textbf{Lightweight BERT Models.} Numerous strategies have been developed to enhance the efficiency of BERT. Techniques like distillation and pre-training followed by fine-tuning are shown to be effective \cite{turc2019well}. Notable methods include BERT-PKD \cite{sun2019patient}, which extracts knowledge from intermediate layers; DistilBERT \cite{sanh2019distilbert}, which uses a triple loss function ; TinyBERT \cite{jiao2020tinybert}, which applies distillation throughout pre-training and task-specific phases; and MobileBERT \cite{sun2020mobilebert}, which introduces a bottleneck structure to reduce parameters. MiniLM \cite{wang2020minilm} further compresses models using deep self-attentive distillation.Building on these advancements, our research integrates both BERT-base and MobileBERT, including their weight-sharing variants, into the supernet.

\textbf{NAS for efficient models.} Recent developments in NAS have significantly enhanced model efficiency using a variety of strategies. For instance, AdaBERT \cite{chen2021adabert} employs differentiable NAS to condense BERT into smaller, task-specific models. Besides, NAS-BERT \cite{xu2021bert} investigates a large supernet with varied architectures, facilitating the creation of models that can be adjusted for size and latency. EfficientBERT  \cite{dong2021efficientbert} uses a three-stage search process to fine-tune MLP configurations within the FFN, which drastically reduces parameters. Notably, AutoBERT-Zero  \cite{gao2022autobert} introduces a novel search space with primitive operators that allow for constructing attention structures and backbones for general pre-trained language models (PLMs) from scratch. To the best of our knowledge, this is the latest NAS method that specifically incorporates lightweight BERT models, which makes it the most pertinent benchmark for our study.

\section{Methodology}

This section outlines the methodology of our proposed NAS method, called ELM. The ELM framework consists of three key components: (i) an expanded search space, (ii) dynamic search strategies for dimensions and head numbers, and (iii) a novel KD technique.







\subsection{Flexible Search Space with High Efficiency}

To enhance the efficiency of our proposed ELM and facilitate the discovery of optimal architectures, we introduce multiple efficient block options into the search space.  This enhanced search space encompasses 12 layers, each providing candidate blocks derived from BERT \cite{devlin2019bert} and MobileBERT \cite{sun2020mobilebert}. These blocks incorporate three weight-sharing mechanisms: no sharing, sharing of the $query$ and $value$, and sharing of the $key$ and $value$, respectively. This configuration allows the construction of a network comprising 12 blocks, where each layer can randomly select from one of six potential combinations.

We adopt the single path one-shot SPOS method \cite{guo2020single} for our NAS approach, constructing the search space as an over-parameterized weight-sharing supernet. The search process comprises two main stages: supernet training and evolutionary search. During training, we uniformly sample paths from $6^{12}$ candidate architectures, focusing optimization efforts solely on the selected blocks. After training, an evolutionary algorithm is employed to sample and evaluate architectures on a validation set. The algorithm iteratively refines the architectural population through selection, crossover, and mutation processes, ultimately identifying the architecture that achieves the highest precision. We maintain a strict parameter ceiling to ensure the exclusion of any architectures that exceed the established constraints.
\begin{figure}[t]
    \centering
    \begin{subfigure}[b]{0.48\linewidth}
        \centering
        \small
        \includegraphics[width=\linewidth]{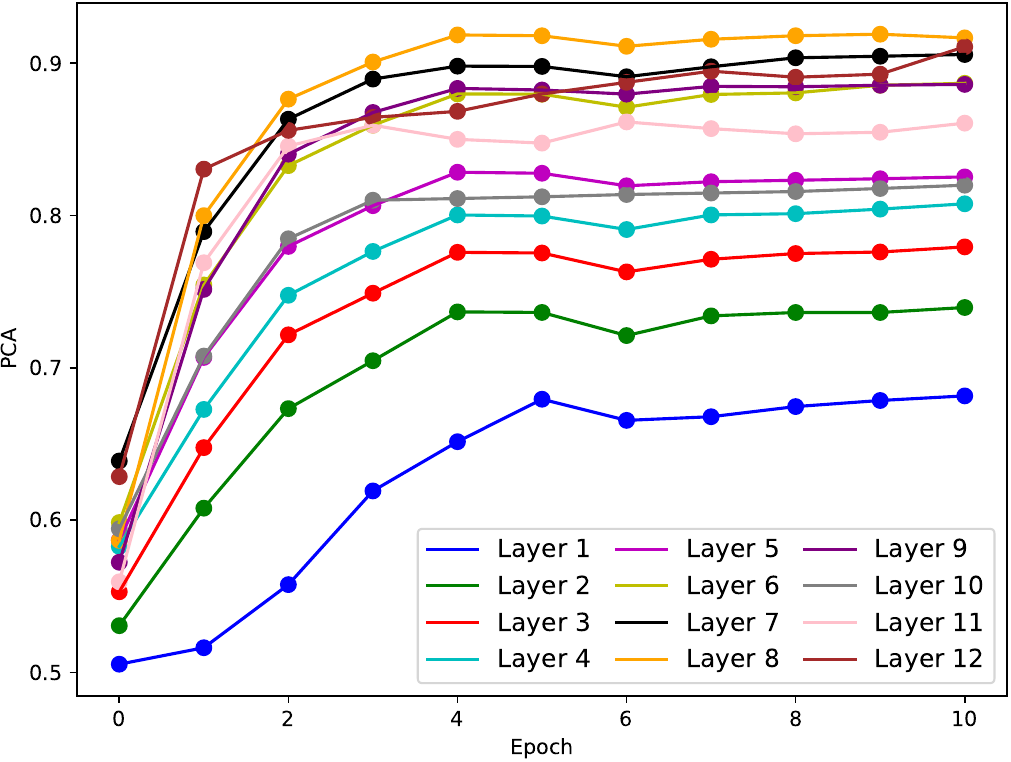}
        \caption{}
    \end{subfigure}
    \hspace{-2mm}
    \begin{subfigure}[b]{0.48\linewidth}
        \centering
        \small
        \includegraphics[width=\linewidth]{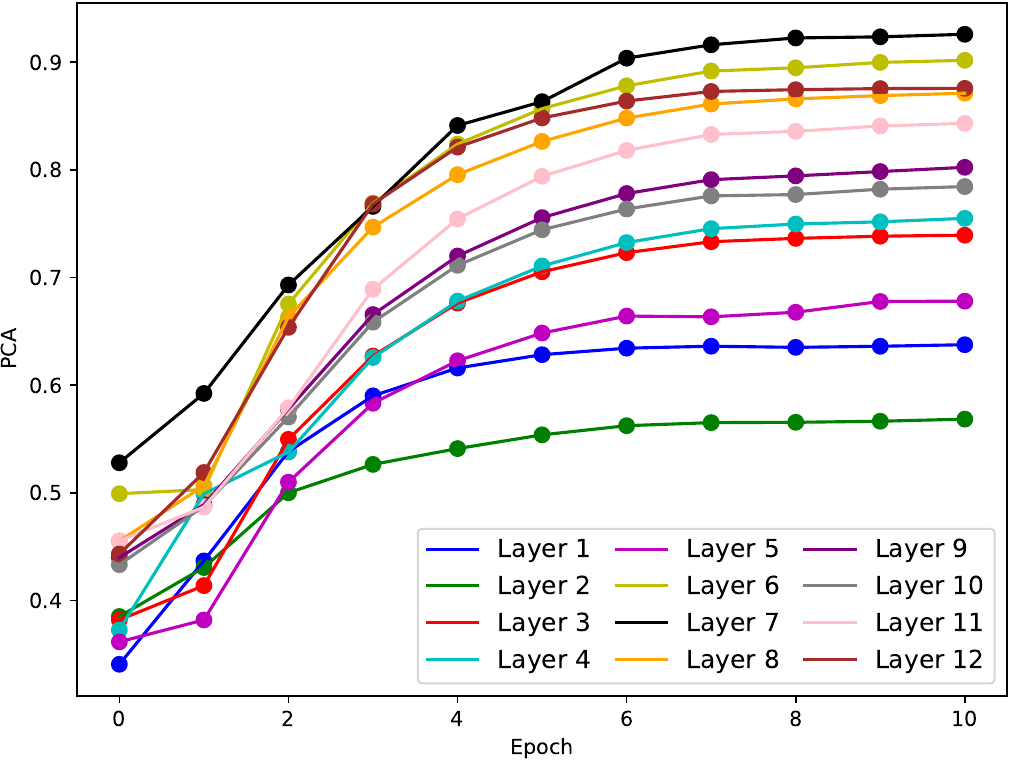}
        \caption{}
    \end{subfigure}
    \vspace{-2mm}
    \caption{Curves of PCA scores of (a) BERT \cite{devlin2019bert} and (b) MobileBERT \cite{sun2020mobilebert} at different epochs during training.}
    \label{fig:pca}
\end{figure}

\begin{figure}[t]
\centering
\begin{subfigure}[b]{0.64\linewidth}
\includegraphics[width=\linewidth, height=0.8\linewidth]{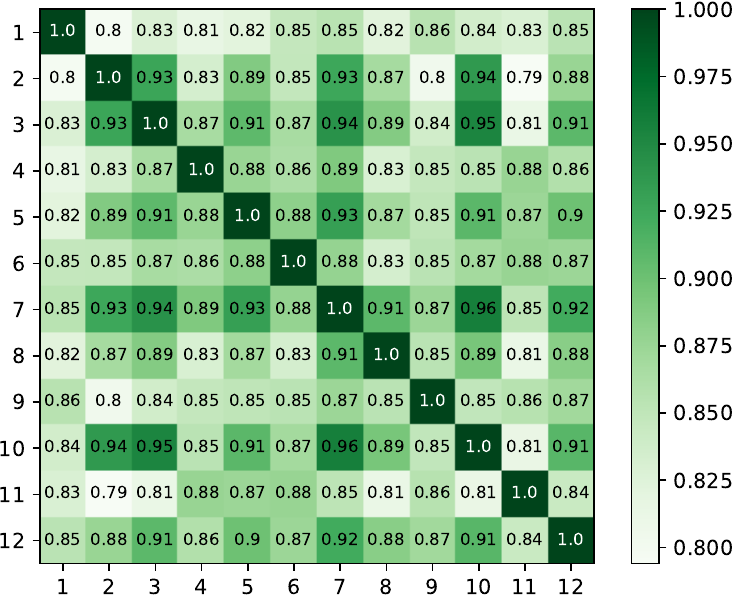}
\caption{}
\end{subfigure}
\begin{subfigure}[b]{0.32\linewidth}
\includegraphics[width=\linewidth, height=1.6\linewidth]{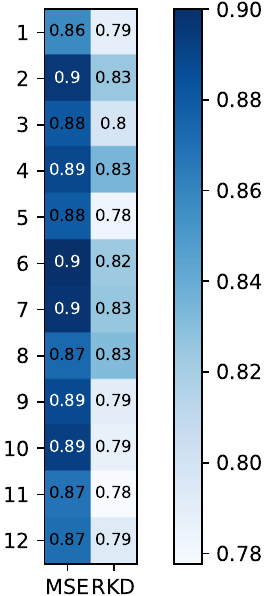}
\caption{}
\end{subfigure}
\vspace{-2mm}
\caption{
(a) CKA comparison among different heads of layer 6 in the searched architecture. (b) average cosine similarity between features of blocks trained with MSE or RKD in each layer.
}
\label{fig:cka_sim}
\end{figure}

\subsection{Efficient Training with Dimension and Head Number Search}

In addition to the 6-block search space, we propose a dynamic dimension adaptation scheme to further enhance the flexibility of the search space. We initially conducted experiments to determine the dimensional preferences of each layer during training by measuring the PCA scores of FFN hidden states over multiple epochs (Fig.~\ref{fig:pca}). The results indicate that (1) PCA scores vary across layers, indicating different optimal dimensions for each, and (2) PCA scores typically increase as training progresses, suggesting that the scores can be used to strategically guide the dynamic allocation of dimensions throughout the training process.

Based on these findings, we initialize the supernet with smaller dimensions and incrementally adjust them based on the PCA scores of the blocks. Specifically, we increase the dimensions of blocks that exhibit the top-$K$ PCA scores until the parameters of the supernet meet the desired constraint. This approach allows for an efficient and effective dimension search by optimizing resource allocation to areas of greatest impact.

\textbf{Searchable head number.} In lightweight models, fixed head counts can lead to inadequately small dimensions per head. For instance, a configuration with a minimum dimension of $192$ divided among $12$ heads results in only $16$ dimensions per head, which is suboptimal. To address this, we introduce a mechanism that allows for the number of heads to be dynamically adjustable.

Rather than creating new independent blocks for each potential head count, we efficiently determine the optimal number of heads by assessing the CKA scores \cite{kornblith2019similarity} between outputs of self-attention heads (Fig.~\ref{fig:cka_sim} (a)). When two heads display high CKA scores, suggesting redundant functionality, one can be removed to allocate more dimensions to the remaining heads.

The procedure for searching the number of heads is outlined in Algorithm~\ref{alg:algorithm2}, where $\eta$ is set to 0.9 during execution. This process ensures a balanced optimization of dimension allocation and model compactness, as summarized in our overall search procedure in Algorithm~\ref{alg:ELM}.

\begin{algorithm}[tb]
\caption{Attention head number search algorithm}
\label{alg:algorithm2}

\begin{algorithmic}[1] 
\REQUIRE Original number of heads $H$, threshold of removing head $\eta$, CKA matrix $Q_l\in\mathbb{R}^{H\times H}$ of layer $l$, dimension of features $C$.
\ENSURE Number of heads preserved in a layer.
\STATE Initialize the set of removed heads $\mathcal{R}\leftarrow \emptyset$
\FOR{$i$ in $1,...,H$}
    \IF{$i$ in $\mathcal{R}$}
        \STATE continue
    \ENDIF
    \FOR{$j$ in $i+1,...,H$}
        \IF {$Q_{i, j} > \eta$}
             \STATE $\mathcal{R}\leftarrow \mathcal{R}\cup \{j\}$
         \ENDIF
     \ENDFOR
 \ENDFOR
 \STATE Searched number of heads $H^* = H - |\mathcal{R}|$
 \STATE \textit{\# Adjust $H^*$ to be evenly divisible.}
 \WHILE{$C\% H^*\neq 0$}
     \STATE $H^*\leftarrow H^*-1$
 \ENDWHILE
 \RETURN Searched head number $H^*$ of layer $l$
 \end{algorithmic}
 \end{algorithm}

\begin{algorithm}[tb]
    \caption{Neural architecture search with ELM}
    \label{alg:ELM}
    \begin{algorithmic}[1]
        \REQUIRE Supernet $\mathcal{S}$, pretrain and downstream datasets $\mathcal{D}_{p}$ and $\mathcal{D}_{d}$, training epochs $E_p$ and $E_d$, and validation dataset $\mathcal{D}_{v}$.
        \STATE \textit{\# (1) Train the supernet $\mathcal{S}$ on pretraining dataset $\mathcal{D}_{p}$}
        \FOR{epoch in $1,...,E_p$}
            \STATE Train supernet $\mathcal{S}$ \& measure PCA scores $P$
            \STATE Select Top-K blocks in $P$ to increase their dimensions
        \ENDFOR
        \STATE \textit{\# (2) Finetune the supernet $\mathcal{S}$ on downstream dataset $\mathcal{D}_{d}$}
        \FOR{epoch in $1,...,E_d$}
            \STATE Train supernet $\mathcal{S}$
        \ENDFOR
        \STATE \textit{\# (3) Search architecture with evolutionary algorithm}
        \STATE $A^* \leftarrow$ EvolutionSearch($\mathcal{S}$, $\mathcal{D}_{v}$)
        \STATE \textit{\# (4) Head number search}
        \FOR{layer $l$ in $1,...,L$}
            \STATE Measure CKA $Q_l$ for layer $l$
            \STATE $H_l^* \leftarrow$ HeadNumberSearch($Q_l$)
        \ENDFOR
        \RETURN Searched architecture $A^*$
    \end{algorithmic}
    
\end{algorithm}

\begin{table*}[t]
    \setlength\tabcolsep{1.5mm}
    \begin{center}
        \caption{Performance comparison of the test set on the GLUE benchmark. Latency measurements of the models are conducted using the NVIDIA A100 GPU.}
        \label{tab:test_glue}
        \vspace{-2mm}
        \small
        \begin{tabular}{@{}l|cc|ccccccccc@{}}
            \toprule
            Model & \#Params & Latency & QNLI & MRPC & SST-2 & CoLA & STS-B & MNLI-m/mm & RTE & QQP & AVG\\
            \midrule
            BERT-base~\cite{devlin2019bert} & 108.9M & 274ms & 90.5 & 88.9 & 93.5 & 52.1 & 85.8 & 84.6/83.4 & 66.4 & 71.2 &79.6 \\
            BERT-base (ours) & 108.9M & 274ms & 91.4 & 88.7 & 93.0 & 49.0 & 87.5 & 84.9/83.9 & 76.6 & 71.3 &80.7  \\
            \midrule
            BERT-tiny \cite{turc2019well} & 14.5M & 44ms & 84.8 & 83.2 & 87.6 & 19.5 & 77.1 & 75.4/74.9 & 62.6 & 66.5 &70.2  \\
            BERT-small \cite{turc2019well} & 28.8M & 79ms & 86.4 & 83.4 & 89.7 & 27.8 & 77.0 & 77.6/77.0 & 61.8 & 68.1 &72.1  \\
            DistilBERT-6 \cite{sanh2019distilbert} & 67.0M & 151ms & 88.9 & 86.9 & \textbf{92.5} & \textbf{49.0} & 81.3 & 82.6/81.3 & 58.4 & 70.1 &76.8  \\
            TinyBERT-4 \cite{jiao2020tinybert} & 14.5M & 45ms & 87.7 & 88.5 & 91.2 & 27.2 & 83.0 & 81.8/80.7 & 64.9 & 69.6 &75.0  \\
            MobileBERT-tiny \cite{sun2020mobilebert} & 15.1M & 62ms & 89.5 & 87.9 & 91.7 & 46.7 & 80.1 & 81.5/81.6 & 65.1 & 68.9 &77.0  \\
            EfficientBERT+ \cite{dong2021efficientbert} & 15.7M & 62ms & 89.3 & \textbf{89.9} & 92.4 & 38.1 & 85.1 & 83.0/82.3 & 69.4 & \textbf{71.2} &77.9  \\
            EfficientBERT++ \cite{dong2021efficientbert} & 16.0M & 65ms & \textbf{90.6} & 88.9 & 92.3 & 42.5 & 83.6 & 83.0/82.5 & 67.8 & \textbf{71.2} &78.0  \\
            \midrule
            ELM-Micro & 5.0M & 15ms & 85.6 & 86.0 & 89.2 & 21.5 & 81.7 & 77.0/76.3 & 66.4& 68.2 &72.4  \\
            ELM-Tiny & 9.7M & 26ms & 88.6 & 87.5 & 91.5 & 30.1 & 84.8 & 81.1/80.2 & 71.7& 70.7 &76.2  \\
            ELM-Small & 15.6M & 48ms & 90.1 & 88.4 & 91.8 & 38.8 & \textbf{86.8} & \textbf{83.2}/\textbf{82.7} & \textbf{75.2}& 70.7 &\textbf{78.6}  \\
            \bottomrule
        \end{tabular}
    \end{center}
\end{table*}

\subsection{Pitfalls of NAS with Knowledge Distillation} \label{sec:kd_loss}

KD has significantly enhanced lightweight BERT models by transferring knowledge from a large teacher model to a smaller student model during training \cite{sanh2019distilbert, jiao2020tinybert}. Recent NAS methods like EfficientBERT  \cite{dong2021efficientbert} and AutoBERT \cite{gao2022autobert} incorporate KD during both the search and retraining phases to maintain consistent objectives and accelerate convergence. This consistency helps bridge the performance gap between the searched architecture and the fully-trained counterpart, which can speed up the architecture search process.

However, we observe that the strict learning objectives of classical KD methods, such as KL divergence and mean squared error (MSE), limit the block diversity within the NAS framework. As indicated in Fig.~\ref{fig:cka_sim}, blocks trained under KD conditions tend to show higher feature similarity compared to those trained independently, which exhibit more pronounced diversity.  This constraint restricts the ability of blocks to express their distinct functional attributes, potentially obscuring standout architectures that could offer unique advantages in the NAS process.

In BERT distillation, the student model is trained to mimic the output of the teacher model using various KD losses. For a training sample $X$, the predicted classification scores are $Y^{(t)}\in\mathbb{R}^{B\times C}$ for the teacher and  $Y^{(s)}\in\mathbb{R}^{B\times C}$ for the student, respectively, where $B$ is the batch size and $C$ is the number of classification categories.

(1) Class distillation loss (KL divergence):
\begin{equation}
    \mathcal{L}_{cd} := \frac{1}{B}\sum_{i=1}^B \sum_{j=1}^C Y_{i,j}^{(t)} \mathrm{log}\left(\frac{Y_{i,j}^{(t)}}{Y_{i,j}^{(s)}}\right),
\end{equation}

Additionally, KD performs the distillation of attention scores  $A_l\in\mathbb{R}^{B,H,N,N}$ and FFN outputs $F_l\in\mathbb{R}^{B,N,C}$ of each layer, where 
$H$ is the number of attention heads and 
$N$ is the token length.

(2) Attention distillation loss:

\begin{equation}
    \mathcal{L}_{ad} = \frac{1}{L}\sum_{l=1}^{L}||A_l^{(t)} - A_l^{(s)}||_2^2 .
\end{equation}

(3) FFN distillation loss:
\begin{equation}
    \mathcal{L}_{fd} = \frac{1}{L}\sum_{l=1}^{L}||F_l^{(t)} - F_l^{(s)}||_2^2 .
\end{equation}

These losses enforce point-to-point feature reconstruction, compelling blocks to produce identical features, thus limiting their capacity to exhibit unique characteristics. Such a rigid alignment between supernet training and individual block training can degrade the ability of NAS algorithms to discern and select optimal architectures effectively.

\textbf{Flexible supernet training with relational losses.} To address this limitation, we propose replacing strict KD losses with relaxed correlation-based losses, inspired by Pearson correlation \cite{huang2022knowledge, cao2022pkd}. These losses focus on capturing the correlation between teacher and student features while being scale-and-shift invariant. The distillation losses are reformulated as follows:

(1) Class distillation loss:
\begin{equation}
    \mathcal{L}_{cd} := \frac{1}{B}\sum_{i=1}^B \left[1 - \rho\left(Y^{(t)}_i, Y^{(s)}_i\right)\right]
\end{equation}

(2) Attention and FFN distillation losses:

\begin{equation}
    \mathcal{L}_{fd} = \frac{1}{L\times B\times N}\sum_{l=1}^{L}\sum_{i=1}^{B\times N}\left[1-\rho\left(F_{l,i}^{(t)}, F_{l,i}^{(s)}\right)\right] .
\end{equation}

By adopting these relaxed correlation-based losses, our approach allows blocks in the supernet to preserve their unique characteristics, which enhances the ability of NAS algorithms in discovering and selecting the suitable architectures.

\begin{table*}[h]
    \setlength\tabcolsep{1.5mm}
    \begin{center}
        \caption{The results of our proposed method are compared with other NAS methods on the GLUE dev set. It is not feasible to make a subcomparison with AutoBERT-Zero-small as it does not provide individual scores for each task in the GLUE dev set.}
        \label{tab:dev_glue}
        \vspace{-2mm}
        \small
        \begin{tabular}{@{}l|c|ccccccccc@{}}
            \toprule
            Model & \#Params &  QNLI & MRPC & SST-2 & CoLA & STS-B & MNLI-m & RTE & QQP & AVG\\
            \midrule
            EfficientBERT-TINY & 9.4M  & 89.3 & 90.1 & 90.1 & 39.1 & 79.9 & 81.7 & 63.2 & 86.7 &77.5  \\
            EfficientBERT & 15.7M & 90.4 & 91.5 & 91.3 & 50.2 & 82.5 & 83.1 & 66.8 & 87.3 &80.4  \\
            AutoBERT-Zero-small & 13.0M & - & - & - & - & - & - & - & - &80.5  \\
            \midrule
            ELM-Micro & 5.0M  & 85.8 & 87.3 & 90.7 & 25.9 & 81.9 & 77.5 & 64.2& 86.9 &75.0  \\
            ELM-Tiny & 9.7M  & 88.9 & 89.3 & 91.4 & 42.4 & 83.7 & 81.2 & 66.4& 87.3 &78.8  \\
            ELM-Small & 15.6M  & \textbf{91.0} & \textbf{92.3} & \textbf{92.9} & \textbf{50.9} & \textbf{85.1} & \textbf{83.8} & \textbf{71.9}& \textbf{88.4} & \textbf{82.0}  \\
            \bottomrule
        \end{tabular}
    \end{center}
\end{table*}

\section{Experiments on Masked Language Modeling}

\begin{table}[t]
    \setlength\tabcolsep{1.5mm}
    \begin{center}
        \caption{Results on SQuAD dev sets. $^*$ represents our implementation.}
        \label{tab:squad}
        \vspace{-2mm}
        \small
        \begin{tabular}{@{}l|c|cccc@{}}
            \toprule
            \multirow{2}{*}{Model} & \multirow{2}{*}{\#Params} & SQuAD v1.1 & SQuAD v2.0 \\
            & & EM/F1 & EM/F1 \\
            \midrule
            BERT-base     & 108.9M  & 80.8/88.5 & -/- \\
            BERT-base$^*$    & 108.9M  & 80.7/88.2 & 75.7/78.7 \\
            \midrule
            TinyBERT-4  & 14.5M  & 72.7/82.1 & 68.2/71.8 \\
            MiniLM-6  & 22.9M & -/- & -/72.7 \\
            EfficientBERT++  & 16.0M  & 78.3/86.5 & 73.0/76.1 \\
            \midrule
            ELM-Micro  & 5.0M  & 70.6/78.2 & 64.8/68.3 \\
            ELM-Tiny  & 9.7M  & 75.2/84.0 & 69.3/73.0 \\
            ELM-Small  & 15.6M  & \textbf{78.5}/\textbf{86.9} & \textbf{73.6}/\textbf{77.1} \\
            \bottomrule
        \end{tabular}
    \end{center}
\end{table}

\subsection{Implementation Details}


To benchmark our model against other lightweight BERT models, we utilize the GLUE \cite{wang2018glue} and SQuAD \cite{rajpurkar2016squad} datasets. We provide the detailed implementation in Appendix \ref{app_imp}.

\subsection{Results on GLUE \& SQuAD}

We summarize our results on GLUE test set, GLUE dev set, and SQuAD in Tables \ref{tab:test_glue}, \ref{tab:dev_glue}, and \ref{tab:squad}, respectively. The results demonstrate that ELM has achieved impressive performance compared with other NAS methods. Specifically, the ELM-Small model, which features a compact design with only 15.6M parameters, surpasses the state-of-the-art EfficientBERT++ model in both accuracy and processing speed. It achieves higher average GLUE scores and outperforms EfficientBERT++ in 4 out of 9 tasks on the GLUE test set. Moreover, on the SQuAD dev set, it surpasses EfficientBERT in terms of EM/F1 metrics for both SQuAD v1.0 and SQuAD v2.0 datasets.

In comparison with other lightweight BERT models, ELM-Small achieves competitive performance with markedly fewer parameters and reduced latency. It also outperforms the AutoBERT-Zero-small model, which has 13M parameters. ELM-Small achieves SOTA performance in each downstream task on the GLUE dev set and attains an average score exceeding that of EfficientBERT. These results demonstrate the effectiveness and efficiency of the proposed ELM approach in generating lightweight BERT models with high performance across various NLU tasks.

We conducted ablation experiments in Appendix  \ref{app_abl}.

\begin{figure}[t]
\centering
\includegraphics[scale=0.6]{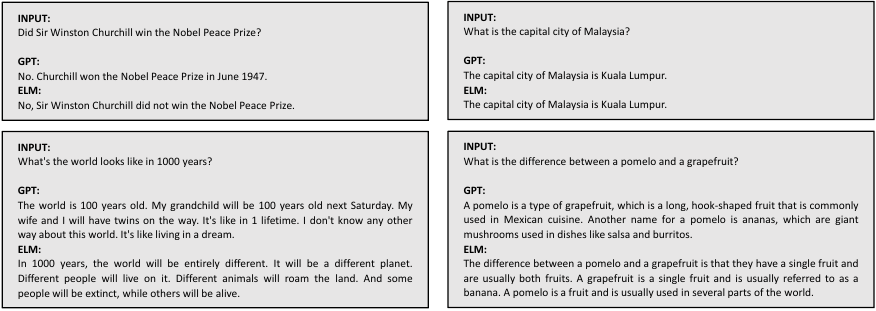}
\begin{center}
\vspace{-2mm}
\captionsetup{singlelinecheck=off}
\caption{Comparison of GPT2-Base and our Chat-ELM-Small trained with MiniLLM on chat task.}
\label{fig:llm_output}
\end{center}
\end{figure}

\begin{table}[t]
    \setlength\tabcolsep{1.5mm}
    \begin{center}
        \caption{Evaluation results referring to the average GPT-4 feedback scores (GPT4) and Rouge-L scores (R-L) obtained from 5 runs.}
        \label{tab:bench}
        \footnotesize
        \renewcommand{\tabcolsep}{0.7mm}
        \vspace{-2mm}
        \begin{tabular}{@{}l|c|cc|cc|cc|c|c@{}}
            \toprule
            \multirow{2}{*}{Model} & \multirow{2}{*}{\#Params} & \multicolumn{2}{c|}{DollyEval}  & \multicolumn{2}{c|}{SelfInst} & \multicolumn{2}{c|}{VicunaEval} & S-NI & UnNI\\
            & & GPT4 & R-L & GPT4 & R-L  & GPT4 & R-L  & R-L & R-L \\
            \midrule
            GPT2     & 120M  & 44.7 & 24.6  & 29.2 & 13.2  & 34.1 & 16.9 & 25.3 & 30.1 \\
            \midrule
            TinyBERT-4& 34.7M  & 37.6 & 19.8 & 26.3 & 10.9  & 29.8 & 13.8 & 20.7 & 25.9 \\
            MobileBERT-tiny & 36.2M  & 39.8 & 20.7 & 27.5 & 11.4  & 32.1 & 14.2 & 21.8 & 26.3 \\
            EfficientBERT++ & 35.7M  & 38.5 & 21.2 & 29.3 & 12.4  & 31.4 & \textbf{15.4} & 22.4 & 26.5 \\
            ELM-Small    & 35M  & \textbf{41.7} & \textbf{23.0} & \textbf{29.6} & \textbf{13.7}  & \textbf{32.2} & 15.1 & \textbf{23.8} & \textbf{28.2} \\
            \bottomrule
        \end{tabular}
    \end{center}
\end{table}

\section{Experiments on Causal Language Modeling}

To validate our performance comprehensively, we further apply our architecture searched on GLUE to the causal language modeling (CLM) task. 

\subsection{Experimental Setup}


\textbf{Pretrain.} We pretrain our model of causal language modeling (CLM) on with 1\% randomly selected texts in SlimPajama~\cite{cerebras2023slimpajama} dataset. We train our ELM-Small with 8 NVIDIA V100 GPUs and a total batch size of 64 for 3 epochs. A cosine annealing learning rate schedule is adopted with an initial learning rate $5\times 10^{-4}$. We use a AdamW optimizer with 0.1 weight decay. Our other setup such as tokenizer and training loss follows GPT2~\cite{radford2019language}.

\textbf{Instruction finetune.} In order to obtain a model for chat application, we finetune our pretrained model with instruction dataset databricks-dolly-15k~\cite{DatabricksBlog2023DollyV2} using KD method MiniLLM~\cite{gu2024minillm}. Specifically, we use GPT2-XL as the teacher model to supervise our Chat-ELM-Small, and all the training data and strategies are the same as MiniLLM.

\textbf{Quantitative evaluation.}

 We conduct numerical evaluation on five instruction-following datasets, including \textbf{DollyEval}, \textbf{SelfInst}~\cite{wang2022self}, \textbf{VicunaEval}~\cite{chiang2023vicuna}, \textbf{S-NI}, and \textbf{UnNI}~\cite{honovich2022unnatural}.

\subsection{Exemplar evaluation} The resulting model Chat-ELM-Small, has $35$M parameters, while obtains competitive performance with the $120$M GPT2-Base. We visualize the chat results of our model and baseline GPT-Base model in Fig.~\ref{fig:llm_output}. 

\subsection{Quantitative Results on Language Benchmarks}


We evaluated the performance of our Chat-ELM-Small model alongside several smaller models trained using MiniLLM~\cite{gu2024minillm}. As shown in Table \ref{tab:bench}, Chat-ELM-Small outperforms the other similarly sized models in 7 out of 8 metrics. Despite having less than one-third the parameters of GPT-2, Chat-ELM-Small still achieves superior results in two metrics on the SelfInst dataset. This highlights the remarkable efficiency of our model, consistently delivering better performance across a variety of language benchmarks while utilizing significantly fewer computational resources.











\section{Conclusion}
This paper presents ELM, a novel NAS method aimed at increasing flexibility in developing compact language models for NLU tasks. ELM enhances the NAS framework by incorporating an expanded search space equipped with dynamic modules and novel knowledge distillation techniques that maintain the unique characteristics of various blocks throughout training. Experimental results show that ELM significantly outperforms existing methods in both masked and causal language modeling tasks, indicating its potential for advancing NLP and deep learning.

\bibliographystyle{IEEEbib}
\bibliography{icme2025references}

\vspace{12pt}

\clearpage
\appendix
\subsection{Implementation Details}
\label{app_imp}

\textbf{Supernet.} In the supernet, each block has a hidden size of 528 and initially contains 12 heads. The FFN starts with an initial hidden dimension of 132, which is a quarter of the hidden size. The hidden dimension has the potential to increase to a maximum of 1056, which is twice the hidden size. After each epoch, dimensionality expansion is applied to the top 21 blocks with the highest PCA values out of the total 72 blocks. This ensures that the average hidden dimension of the FFN closely matches the hidden size after training. The inner hidden size of the MobileBERT series blocks consistently remains at 132.

\textbf{Training.} Our training strategy and datasets mostly follow EfficientBERT \cite{dong2021efficientbert}. In the NAS training, we pretrain the supernet on English Wikipedia \cite{devlin2019bert} and BooksCorpus \cite{zhu2015aligning}, then utilize 90\% of the training set from each GLUE task for fine-tuning. We reserve the remaining 10\% of the MNLI task to evaluate the accuracy of architectures in search.
After obtaining the searched architecture, we pretrain the model on
the complete English Wikipedia and BooksCorpus.

The number of training epochs is set to 10 for all tasks, except for the CoLA task, which undergoes fine-tuning for 50 epochs. During the fine-tuning stage of the downstream task model, the batch size is set to 32, while for the other stages, it is set to 256. For the two pre-training stages, the learning rate is set to 1e-4, while for the first fine-tuning stage, the learning rate is adjusted to 4e-4. Fine-tuning the GLUE dataset is done with a fixed learning rate of 5e-5, On the other hand, the learning rate for fine-tuning the SQuAD dataset is set to 1e-4. The training process utilizes Adam as the optimizer, with $\beta_{1}=0.9$ and $\beta_{2}=0.999$ being used. The weight decay is set to 0.01. A linearly-annealing learning rate is adopted with a warm-up ratio of 0.1.

\textbf{Searching.} The SPOS algorithm sets the population size to 50 and the number of generations to 40 in the genetic algorithm. The crossover probability is set to 1, and the mutation probability is set to 0.1. We use parameters constraints of $15.7$M, $10$M, and $5$M for ELM-Small, ELM-Tiny, and ELM-Micro, respectively. Specifically, for ELM-Micro, the hidden dimension and number of layer are reduced to $192$ and $6$, respectively.

\subsection{Ablation Study}
\label{app_abl}
We conduct experiments on GLUE dev set to validate the efficacy of the proposed modules. 

\textbf{Dynamic dimensions.} The results shown in Table \ref{ab_fix} are obtained by setting the hidden dimensions of each FFN model to align with the input dimensions. Dynamic dimensions effectively improve model performance, especially in smaller models.

\textbf{Head number searching.} The research findings presented in Table \ref{ab_heads} demonstrate that the omission of attention head search from both the smaller and larger models results in a reduction in GLUE scores, specifically in the case of the Micro model. This suggests that, with fewer dimensions per head, optimizing the number of attention heads becomes crucial for maintaining model accuracy.

\begin{table}[t]
    \centering
    \begin{minipage}[t]{0.47\linewidth}
        \centering
        \captionof{table}{Effect of our dimension search. \textsuperscript{$\dagger$} represents predefined and fixed dimension.}
        \label{ab_fix}
        \small
        \vspace{-2mm}
        \begin{tabular}{@{}l|c|c@{}}
            \toprule
            Size & \#Params   & GLUE\\
            \midrule
            Micro & 5.0M & 75.0  \\     Micro\textsuperscript{$\dagger$} & 5.0M &74.2  \\
            \midrule
            Small & 15.6M  &82.0  \\    Small\textsuperscript{$\dagger$} & 15.6M  &81.5  \\
            \bottomrule
        \end{tabular}
    \end{minipage}
    \hfill
    \begin{minipage}[t]{0.47\linewidth}
        \centering
        \caption{Effect of attention head search (AHS). \textsuperscript{$\dagger$} represents the models without AHS.}
        \label{ab_heads}
        \small
        \vspace{-2mm}
        \begin{tabular}{@{}l|c|c@{}}
            \toprule
            Size & \#Params   & GLUE\\
            \midrule
            Micro & 5.0M & 75.0  \\     Micro\textsuperscript{$\dagger$} & 5.0M &74.2  \\
            \midrule
            Small & 15.6M  &82.0  \\        Small\textsuperscript{$\dagger$} & 15.6M  &81.7  \\
            \bottomrule
        \end{tabular}

    \end{minipage}
\end{table}

\begin{table}[t]
    \centering
    \begin{minipage}[t]{0.47\linewidth}
        \centering
        \caption{Effect of our relational KD losses. \textsuperscript{$\dagger$} represents original losses.}
        \label{ab_mse}
        \small
        \vspace{-2mm}
        \begin{tabular}{@{}l|c|c@{}}
            \toprule
            Size & \#Params   & GLUE\\
            \midrule
            Micro & 5.0M & 75.0  \\     Micro\textsuperscript{$\dagger$} & 4.9M &73.8  \\
            \midrule
            Small & 15.6M  &82.0  \\    Small\textsuperscript{$\dagger$} & 15.6M  &81.2  \\
            \bottomrule
        \end{tabular}
        
    \end{minipage}
    \hfill
    \begin{minipage}[t]{0.47\linewidth}
        \centering
        \captionof{table}{Effect of weight sharing blocks in search space. \textsuperscript{$\dagger$} represents no weight sharing.}
        \label{ab_01}
        \small
        \vspace{-2mm}
        \begin{tabular}{@{}l|c|c@{}}
            \toprule
            Size & \#Params   & GLUE\\
            \midrule
            Micro & 5.0M & 75.0  \\     Micro\textsuperscript{$\dagger$} & 5.0M &74.3  \\
            \midrule
            Small & 15.6M  &82.0  \\    Small\textsuperscript{$\dagger$} & 15.7M  &81.6  \\
            \bottomrule
        \end{tabular}
        
    \end{minipage}
\end{table}

\textbf{New KD losses.} Table \ref{ab_mse} demonstrates a notable improvement in performance with our new KD losses compared to MSE losses, especially for smaller models. Experimental results indicate that the relaxation of the loss function is highly advantageous for our NAS algorithm in discovering improved network structures.

\textbf{Weight sharing blocks.} By removing the weight sharing blocks from the supernet (leaving only 2 blocks per layer) and reducing the number of blocks added per epoch to one-third of the original, the results shown in Table \ref{ab_01} are obtained. he results indicate that incorporating weight-sharing blocks enhances the discovery of more compact and efficient network architectures.

\begin{figure*}[t]
\centering
\includegraphics[scale=0.66]{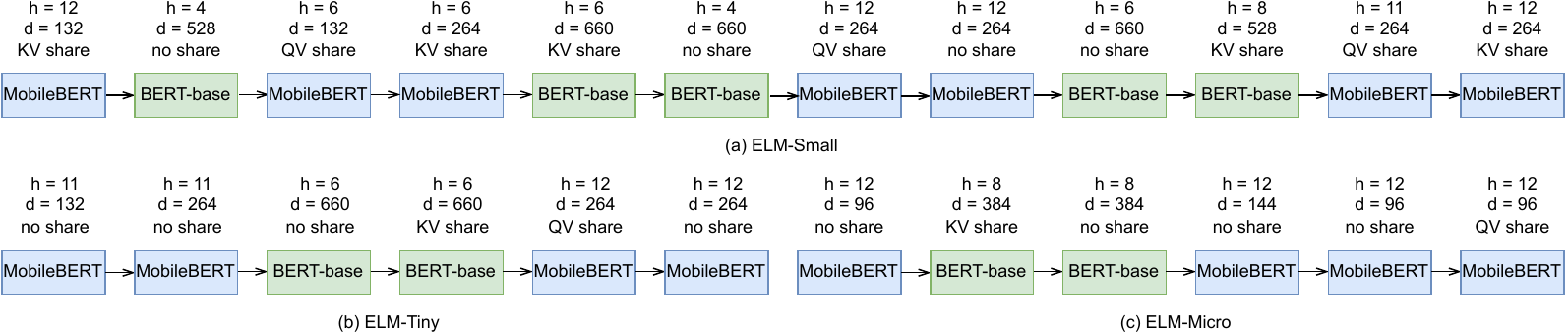}
\begin{center}
\vspace{-2mm}
\captionsetup{singlelinecheck=off}
\caption{The individual architectures of ELM-Small/Tiny/Micro are displayed from top to bottom. H represents the number of heads and d represents the hidden dimensions, while QV/KV/no share indicate the different ways of weight sharing.}
\label{fig:arch}
\end{center}
\end{figure*}
\textbf{Visualization of searched architectures.} Fig.~\ref{fig:arch} illustrates the searched architectures of ELM-Small/Tiny/Micro. We can infer that, the searched models prefer more dimensions in the middle stage of layers; while in the first layer, the dimension is small since it has fewer semantic information than the posterior layers. More MobileBERT blocks are involved compared to BERT blocks, as it achieves better efficiency-performance balance.

\begin{figure}[t]
    \centering
    \begin{subfigure}[b]{0.48\linewidth}
        \centering
        \small
        \includegraphics[width=0.94\linewidth]{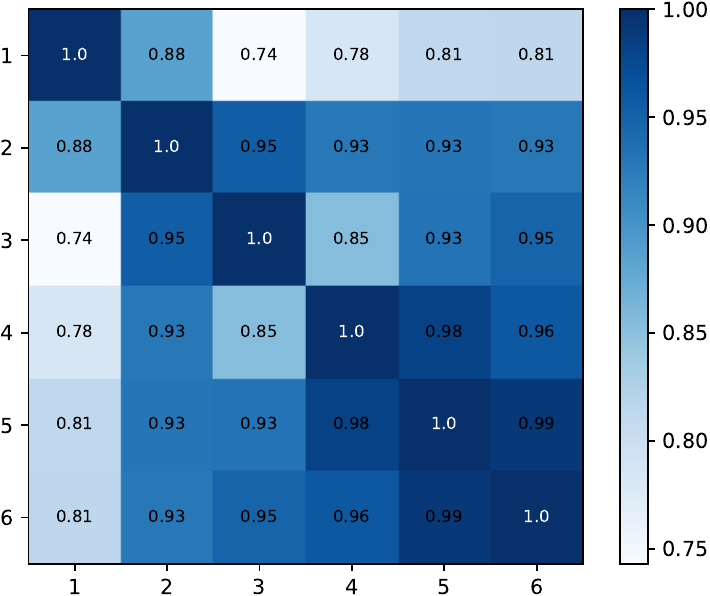}
        \caption{}
    \end{subfigure}
    \hspace{-2mm}
    \begin{subfigure}[b]{0.48\linewidth}
        \centering
        \small
        \includegraphics[width=0.94\linewidth]{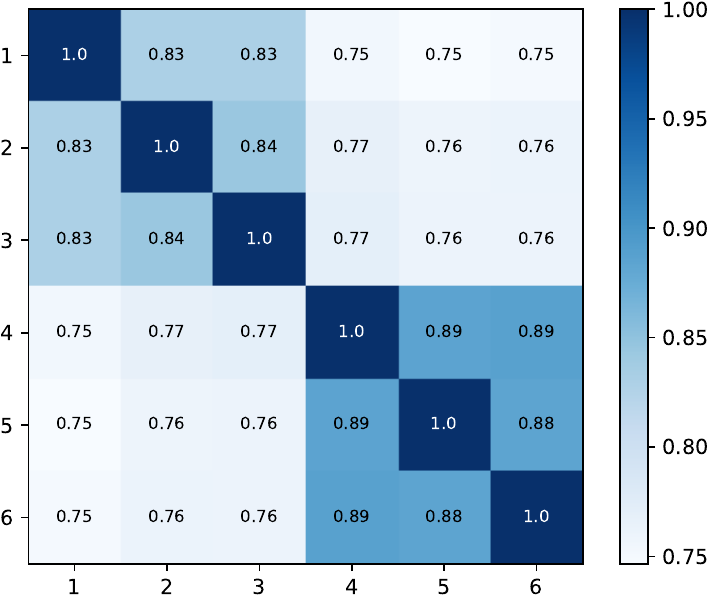}
        \caption{}
    \end{subfigure}
    \caption{Taking layer 6 as an example, the similarity between blocks is compared using (a) MSE and (b) RKD loss functions. Blocks with numbers 0 to 2 belong to the BERT-base series, while blocks with numbers 3 to 5 belong to the MobileBERT series. The numbers obtained by taking the modulo of 3 represent no sharing, QV sharing, and KV sharing, respectively.}
    \label{fig:sim}
\end{figure}

\textbf{Similarities between blocks.} Fig.~\ref{fig:sim} presents a comparison of the similarity distributions among blocks when employing MSE and RKD loss functions, with layer 6 of ELM-Small supernet as an illustrative example. Clearly, the utilization of the RKD loss function leads to diminished similarity values between blocks, thereby yielding a heightened training effect for the layer.

\textbf{Efficiency analysis.} It has been empirically demonstrated that employing dynamic dimensions enhances algorithmic efficiency. The ELM-Small model required 7.1 GPU days to complete execution, whereas utilizing fixed dimension search required 8.5 GPU days. In general, our algorithmic efficiency surpasses that of EfficientBERT. Reproducing the search procedure of the equivalently-sized EfficientBERT model in the identical experimental setting necessitated 9.2 GPU days.

\subsection{Code Availability} We released it on https://github.com/ra225/ELM

\end{document}